\documentclass{article} 
\usepackage{iclr2019_conference,times}

\usepackage{hyperref}
\usepackage{url}

\usepackage[T1]{fontenc}
\usepackage[latin1]{inputenc}
\usepackage{subeqnarray}
\usepackage{amsfonts}
\usepackage{amsmath}
\usepackage{amssymb}
\usepackage{graphicx}
\usepackage{color}
\usepackage{stfloats}
\usepackage{balance}
\usepackage{epstopdf}
\usepackage{minibox}
\usepackage{xcolor,colortbl}
\usepackage{algpseudocode}
\usepackage{multirow}
\usepackage{hhline}

\usepackage[numbers]{natbib}

\title{Estimating skeleton-based gait abnormality index by sparse deep auto-encoder}

\author{Trong-Nguyen Nguyen\\
DIRO, University of Montreal\\
Montreal, QC, Canada\\
\texttt{nguyetn@iro.umontreal.ca}\\
\And
Huu-Hung Huynh\\
University of Science and Technology\\
Danang, Vietnam\\
\texttt{hhhung@dut.udn.vn}\\
\And
Jean Meunier\\
DIRO, University of Montreal\\
Montreal, QC, Canada\\
\texttt{meunier@iro.umontreal.ca}\\
}

\iclrfinalcopy 
\begin{document}

\maketitle

\begin{abstract}
This paper proposes an approach estimating a gait abnormality index based on skeletal information provided by a depth camera. Differently from related works where the extraction of hand-crafted features is required to describe gait characteristics, our method automatically performs that stage with the support of a deep auto-encoder. In order to get visually interpretable features, we embedded a constraint of sparsity into the model. Similarly to most gait-related studies, the temporal factor is also considered as a post-processing in our system. This method provided promising results when experimenting on a dataset containing nearly one hundred thousand skeleton samples.
\end{abstract}

\section{Introduction}
Walking gait is an important factor to assess patient's health. Many studies have been proposed to deal with the task of gait analysis using various input data types such as silhouette, depth map, skeleton, or even raw color image. In this work, the skeleton is employed to represent the input posture because it contains 3D body joint positions to directly analyze gait while the others need further processing to produce appropriate gait-related information.

Some recent skeleton-based approaches work on a similar scheme where the posture corresponding to each frame is represented by hand-crafted features, and a model can embed the temporal factor to learn the sequential gaits as well as to estimate a likelihood for each input gait sequence.~\citeauthor{Paiement2014}~\cite{Paiement2014,Tao2016} proposed statistical models to estimate the quality of human movement based on a sequence of skeletons captured by a Kinect. The model was formed according to patterns of normal walking gait. The assessment could be performed on each individual skeleton as well as a sequence of frames since the hidden Markov model (HMM) was used to represent the change of consecutive postures. The study~\cite{Nguyen2016} also employed a HMM to model the walking movement. Differently from the two mentioned works, the researchers~\cite{Nguyen2016} considered the task they were dealing with as an unsupervised learning problem while the others were supervised. There are two main differences between our method and those studies. First, we do not use hand-crafted features (such as positions, velocities, distances or pairwise joint angles as in~\cite{Paiement2014,Tao2016,Nguyen2016}) to describe a posture. This task is automatically performed without our supervision using a deep auto-encoder. The resulting features may not be interpretable under geometrical aspect but they are still useful as showed in our experimental results (see Section~\ref{sec:results}). Second, we believe that the normality/abnormality of a walking gait is contained inside each individual posture itself. Therefore, our model estimates the abnormality index on each skeleton instead of a sequence. In other words, the temporal factor is not integrated into our model but as a post-processing. However, it still plays an important role in the assessment since it can smooth a noisy sequence of per-skeleton abnormality indices. 

The remainder of this paper is organized as follows: the model structure and the gait assessment scheme are presented in Section~\ref{sec:method}; Section~\ref{sec:experiments} describes the dataset and results of our experiments; and the conclusion is given in Section~\ref{sec:conclusion}.

\section{Proposed method} \label{sec:method}
In our work, the skeletons are determined based on the depth maps captured by a Kinect 2~\cite{Shotton2011realtime,Shotton2013efficient}. Each skeleton is represented by a collection of 25 joint positions in 3D space.

\subsection{Joint selection and coordinate normalization} \label{sec:preprocessing}
\begin{figure}[t]
\centering
\scalebox{1}{
\begin{picture}(160,190)
	\put(0,0){\includegraphics[scale=1]{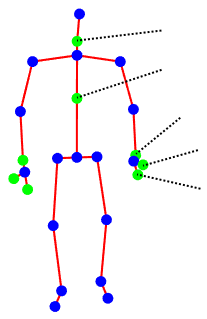}}
	\put(102,172){neck}
	\put(102,150){mid spine}
	\put(113,121){wrist}
	\put(125,100){thumb}
	\put(127,75){hand tip}
\end{picture}}
\caption{Joint selection in our work. Blue joints are selected and green joints are discarded. The input skeleton contains 25 joints that are provided by a Kinect 2.}
\label{fig:skeleton}
\end{figure}
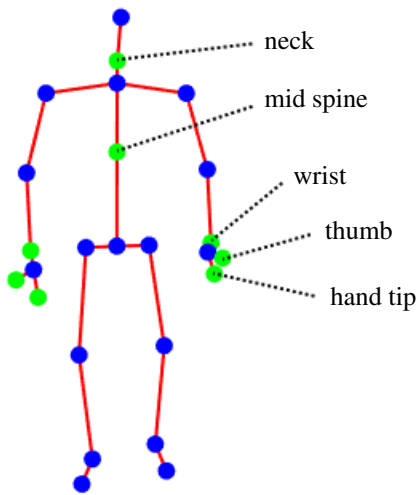
Before feeding skeleton data into our model, a pre-processing is performed in order to reduce the effect of less important joints as well as to normalize the data range. Concretely, we discard 8 joints including neck, hand tips, thumbs, mid spine and wrists. The neck is removed because it can be interpolated according to the head and the shoulder spine, while the collection of hand tips, thumbs and wrists is related to the hand joints. The mid spine is also discarded because its motion is significantly limited compared with the remaining joints. A visualization of the selected and discarded joints is shown in Fig.~\ref{fig:skeleton}.

Since the position of a joint is a point in 3D space, we separate the input joint coordinates into 3 groups corresponding to 3 axes. In other words, they can be considered as the projections of a skeleton onto the X-, Y- and Z-axes. Therefore, each skeleton is represented by 3 vectors of size 17.

The next step that enhances the input data for our model is data range normalization. This task is independently performed on each of the 3 possible inputs corresponding to a skeleton. Since each input is the joint coordinates along an axis, we simply scale its range into [0, 1]. This normalization synchronizes the input data range while the pair-wise distance relation of skeletal joints along each axis is still kept.

\subsection{Auto-encoder structure} \label{sec:structure}
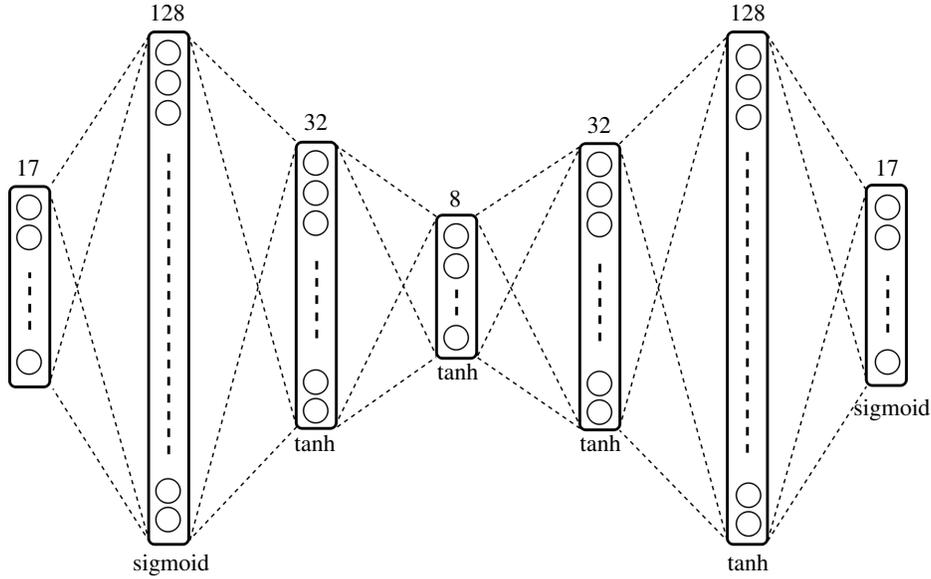
\begin{figure*}[t]
\centering
\scalebox{0.9}{
\begin{picture}(390,235)
	\put(0,10){\includegraphics[scale=0.333]{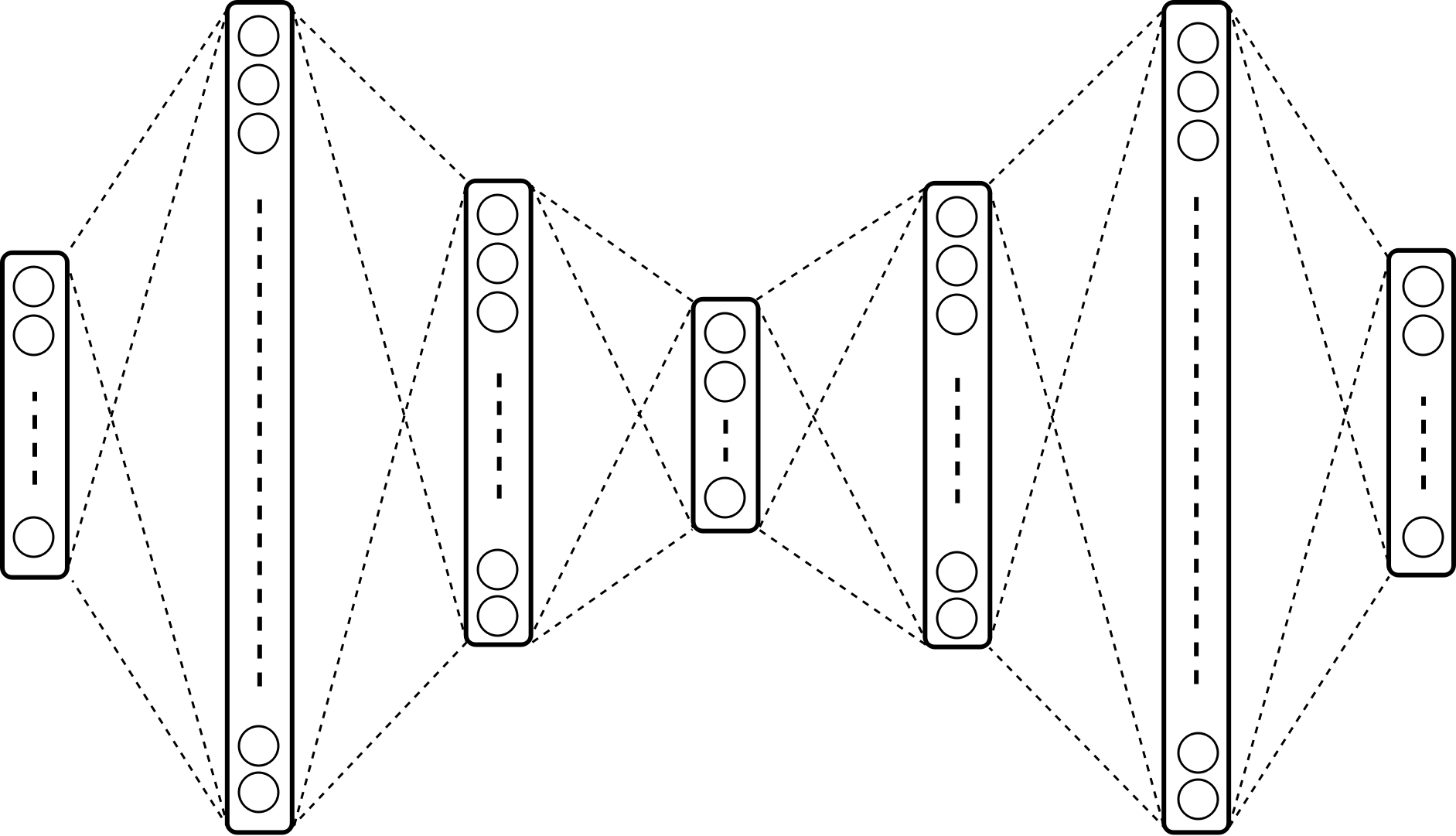}}
	\put(3,166){17}\put(364,166){17}
	\put(59,231){128}\put(303,231){128}
	\put(124,185){32}\put(243,184){32}
	\put(185,153){8}
	\put(52,0){sigmoid}\put(302,0){tanh}
	\put(120,50){tanh}\put(240,50){tanh}
	\put(180,80){tanh}\put(355,65){sigmoid}
\end{picture}}
\caption{The structure of our proposed auto-encoder. Each layer is fully connected to the next one. The number of units in each layer is shown at the top and the corresponding activation function is presented at the bottom. The middle layer represents the latent space. Notice that we use 3 independent auto-encoders for 3 inputs (i.e. X-axis, Y-axis and Z-axis data).}
\label{fig:structure}
\end{figure*}
The use of a deep auto-encoder in our work aims to two purposes. First, the middle layer of an auto-encoder can be considered as a feature space (so-called latent space) since it highlights useful information embedded in the model's input. Second, an auto-encoder is a stack of an encoder and a decoder, this is thus appropriate for our work since the two components can learn shared properties of the training data. Another reason of employing an auto-encoder is that we consider this work as an unsupervised learning problem. A sketch of our model is shown in Fig.~\ref{fig:structure}. In the training stage, the output units are assigned the same values as the input. In practice, the model attempts to reconstruct an output that is similar to its input.

A noticeable factor in our model is the large number of units in the second layer. The conversion of data dimension from 17 (the input layer) to 128 (the second layer) may lead the input to a redundant representation. However, by adding a sparsity constraint into the optimization objective, we expect to determine interesting structures of input skeletons. Concretely, each hidden unit in the second layer focuses on determining a particular property of the input. A hidden unit in the third layer can be considered as a weighted combination of raw features determined in the previous one, and so on for the middle layer. The remaining layers, i.e. the decoder except the middle layer, are simply formed as a reflection of the encoder since auto-encoders usually have a symmetric structure.

We employed two activation functions in our model: sigmoid and tanh. Since the tanh function has some advantages compared with sigmoid~\cite{cs231n}, we assign it to two-third of the layers. The use of sigmoid at the output layer is to force the output range being asymptotically similar to the input. In other words, the choice of output activation depends on the way that input normalization has been performed. The sigmoid is also employed in the second layer because of the sparsity constraint. Concretely, this penalty term is defined as a sum of Kullback-Leibler divergences between variables $\rho$ and $\hat{\rho}_j$:
\begin{equation}
	 P = \sum_{j=1}^{128} \rho \mathrm{log} \frac{\rho}{\hat{\rho}_j} + (1 - \rho) \mathrm{log} \frac{1-\rho}{1-\hat{\rho}_j} 
\label{eq:sparsity}
\end{equation}
where $\rho$ is a small value that is close to zero ($\rho=0.05$ in our experiments) and $\hat{\rho}_j$ is the average activation (calculated over a training batch) of the $j^{th}$ unit in the second layer. By assigning a sigmoid activation to the second layer, we can guarantee that $\hat{\rho}_j$ is always positive, and the logarithm thus easily performs.

In addition, a L2-regularization term is also added into the optimization objective to reduce the risk of overfitting. Therefore, the loss function of our auto-encoder includes the reconstruction loss (mean square error between model input and output), the KL-divergence penalty, and the L2-regularization. Let us notice that we use 3 models for 3 input types (i.e. X-axis, Y-axis and Z-axis data) and they are independently trained.

\subsection{Gait abnormality index estimation} \label{sec:assessment}
Since our model is unsupervised, the training stage is performed with skeletons belonging only to the normal walking gait category. The auto-encoder is expected to modify its parameters according to shared properties of such inputs. Therefore, input of an abnormal walking gait (rather than normal ones) should result in a bad reconstruction with a high error.

Recall that each skeleton provides 3 inputs for 3 auto-encoders, a simple summation (or average) of errors thus seems to be reasonable as a gait abnormality index. However, data along different axes should have different contributions to the index. Therefore, we decide to estimate the index as a weighted sum of reconstruction errors, in which the weights are measured according to the errors measured on the training data. Concretely, we calculate the MSE of training data according to the 3 models at the end of the training stage. The weight corresponding to a model $k$ with $k\in \{\mathrm{X},\mathrm{Y},\mathrm{Z}\}$ is defined as
\begin{equation}
	w_k = e_k^{-1}\sum_ke_k
\label{eq:weight}
\end{equation}
where $e_k$ is the MSE computed over the training set of model $k$. This is a reasonable definition since a model that returns a low error should has a high weight. Besides, eq.~(\ref{eq:weight}) guarantees the relation between the ratio of errors and the ratio of weights between a pair of models. Although using only $e_k^{-1}$ also satisfies this relation, the summation of training errors allows the returned weight to lie in a reasonable range.

The gait abnormality index of a skeleton is thus estimated as a weighted sum of reconstruction errors resulting from the 3 auto-encoders. In order to estimate this quantity for a sequence of skeletons, we simply compute the mean value. As presented in Section~\ref{sec:results}, this calculation provides a good measurement of gait index since it reduces the effect of possible noisy postures in the sequence.

\section{Experiments} \label{sec:experiments}
Gait normality/abnormality index is a term that has been differently defined in various studies, there is thus no ground-truth index for an approach assessment. Therefore, we performed the experiments according to a popular application of gait abnormality index estimation. In most related studies, such index is used to categorize a walking gait into one of the two groups: normal and abnormal gaits. This section focuses on the ability of our approach in estimating an index that easily classifies an input gait.

\subsection{Dataset} \label{sec:dataset}
The dataset used in our experiments includes normal gaits and 8 abnormal gaits that were performed on a treadmill by 9 volunteers using a Kinect 2 that determined subject skeletons under frontal view. The abnormal walking gaits were simulated by (a) padding a sole of 5, 10 or 15 centimeters under one foot, or (b) attaching a 4 kilograms weight to one ankle. Each gait was acquired as a sequence of 1200 consecutive skeletons. The treadmill speed was 1.28 kph. In addition to skeletons, we also recorded the corresponding sequence of 2D silhouettes in order to perform a comparison with another related work that requires such data. 

The dataset was split into two sets, in which the training set consisted of gaits performed by 5 subjects and the gaits of the 4 remaining volunteers were employed to evaluate our trained models. That split was also the same when we experimented other related works to provide a comparison.

\subsection{Experimental results} \label{sec:results}
\begin{figure}[t]
\centering
	\includegraphics[width=0.6\textwidth]{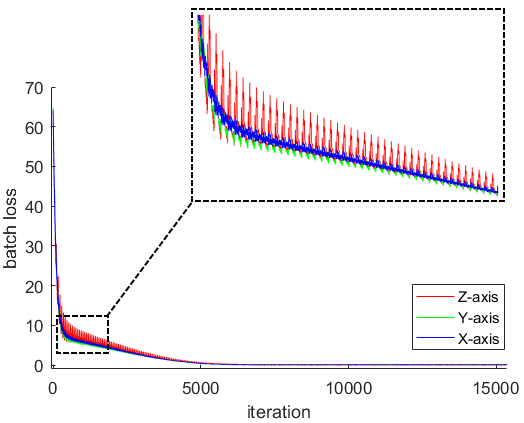}
\caption{Batch losses of 3 auto-encoders during the training stage. The loss was computed according to reconstruction loss, KL-divergence penalty, and L2-regularization as described in Section~\ref{sec:structure}.}
\label{fig:batchloss}
\end{figure}

Each auto-encoder was trained with 6000 samples corresponding to 5 sequences of normal walking skeletons in the training set. The batch losses during the training stage is shown in Fig.~\ref{fig:batchloss}. It is obvious that the batch loss resulting from the X- and Y-models is much more stable than the Z-model.

Since we are dealing with a one-class classification problem where the decision normal vs. abnormal depends on a threshold on the reconstruction error, evaluating the approach according to only the classification accuracy is not enough. The Area Under Curve (AUC) of the Receiver Operating Characteristic (ROC) curve is chosen as the measure of our model ability. This quantity has been also employed in many studies related to binary decision. According to the ROC curve, we estimated the Equal Error Rate (EER) and then calculated other evaluation quantities including sensitivity, specificity, precision, accuracy, and F1-score. Our experimental results are shown in Table~\ref{table:result}. The per-frame index indicates the estimation based on each individual skeleton. The per-segment index was computed as the average per-frame index over a short sequence of skeletons (length of 20 in our experiments). The per-sequence index was the average index over the entire sequence, i.e. 1200 consecutive skeletons for each gait of a subject. 
\begin{table*}[t]
\centering
\caption{Experimental results of our proposed approach (from coarse to fine).}
\label{table:result}
\vspace{5pt}
\scriptsize
\begin{tabular}{l||ccccccc}
\hline
\textbf{Index estimation} & \textbf{AUC} & \textbf{EER} & \textbf{Sensitivity} & \textbf{Specificity} & \textbf{Precision} & \textbf{Accuracy} & \textbf{F1-score} \\ \hline \hline
X-axis model & 0.818 & 0.266 & 0.734 & 0.734 & 0.957 & 0.734 & 0.831 \\
Y-axis model & 0.762 & 0.303 & 0.697 & 0.697 & 0.948 & 0.697 & 0.803 \\
Z-axis model & 0.619 & 0.419 & 0.581 & 0.581 & 0.917 & 0.581 & 0.711 \\ \hline 
per-frame sum & 0.748 & 0.316 & 0.684 & 0.684 & 0.945 & 0.684 & 0.794\\
per-segment sum & 0.812 & 0.284 & 0.716 & 0.717 & 0.953 & 0.716 & 0.818\\
per-sequence sum & 0.844 & 0.278 & 0.719 & 0.750 & 0.958 & 0.722 & 0.821\\ \hline 
per-frame weighted sum & 0.855 & 0.230 & 0.770 & 0.771 & 0.964 & 0.770 & 0.856 \\ 
per-segment weighted sum & 0.910 & 0.188 & 0.812 & 0.813 & 0.972 & 0.812 & 0.885 \\ 
per-sequence weighted sum & \textbf{0.945} & \textbf{0.139} & \textbf{0.844} & \textbf{1.000} & \textbf{1.000} & \textbf{0.861} & \textbf{0.915} \\ \hline
\end{tabular}

\end{table*}
\begin{figure*}[t]
\centering
\begin{picture}(520,280)
	\put(0,15){\includegraphics[scale=1.25]{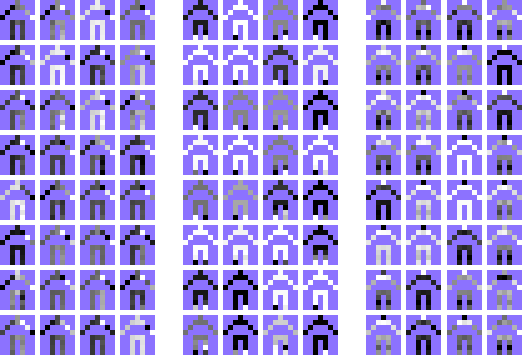}}
	\put(12,0){(a) X-axis hidden units}
	\put(150,0){(b) Y-axis hidden units}
	\put(288,0){(c) Z-axis hidden units}
\end{picture}
\caption{Visualization of some hidden units (that can be considered as filters) in the second layer of the 3 auto-encoders. Each unit is represented by a collection of 17 joint's weights (gray pixels placed as a skeleton). The brighter a pixel is, the higher corresponding weight is. Recall that the Kinect coordinate system is right-handed and the origin is at the sensor. The X-, Y- and Z-axes correspond to leftward, upward and forward directions, respectively.}
\label{fig:maps}
\end{figure*}
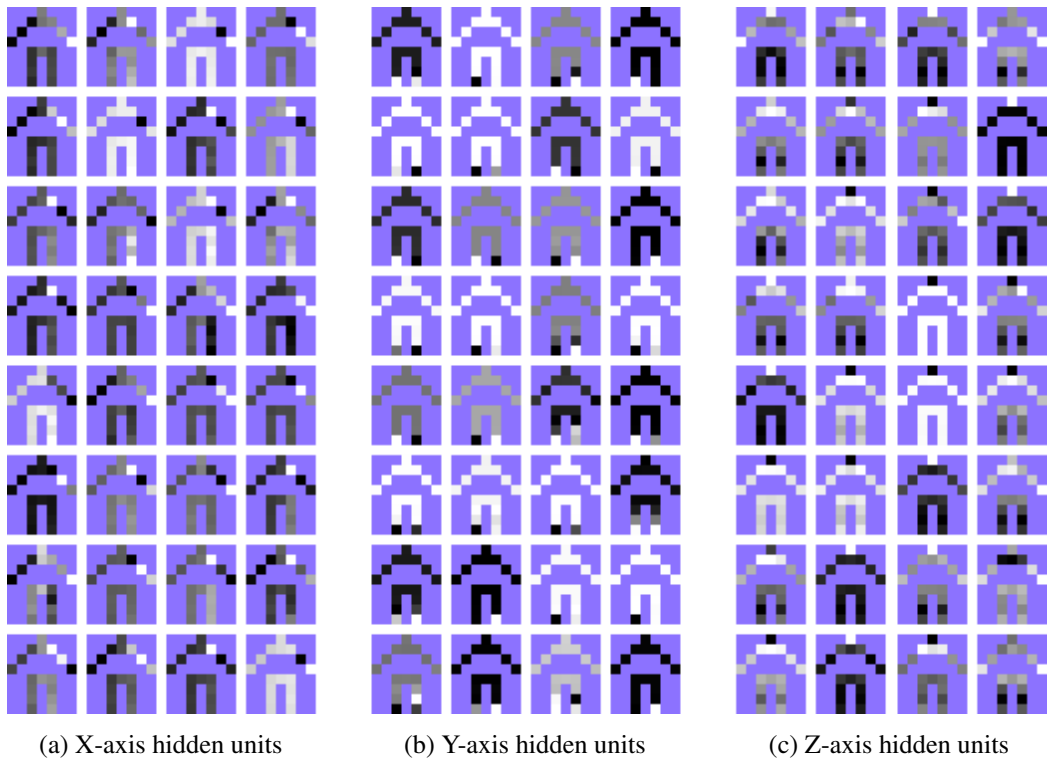

According to Table~\ref{table:result}, the indices estimated by the X-axis and Y-axis auto-encoders were significantly better than the Z-axis model. The batch loss stability of the three models in Fig.~\ref{fig:batchloss} has reflected this property. Table~\ref{table:result} also showed that an abnormality index computed as a weighted sum of elementary indices provided best results in distinguishing between normal and abnormal walking gaits. When embedding the temporal factor into the index estimation, our approach provided the best results whichever using weighted or non-weighted sums. It confirmed that the temporal movement always plays an important role in gait analysis.

In order to get an overall understanding about the raw features in our auto-encoders, we visualize some hidden units in the second layer of each model in Fig.~\ref{fig:maps}. As mentioned in Section~\ref{sec:preprocessing}, the input of each model is the joint coordinates projected onto an axis. According to Fig.~\ref{fig:maps}(a), raw features of the X-axis model tended to measure joint differences along the horizontal orientation since the X-axis corresponds to the leftward direction. In the same fashion, raw features of the model of Y-axis (upward direction) focused on the lower body versus the upper one. The Z-axis filters were more difficult to interpret since they seemed noisy. This may be due to the large movement of body joints along the forward axis, e.g. shoulder joints were salient in many filters in Fig.~\ref{fig:maps}(c) and their movement was significantly less than the others. 

\subsection{Related-works comparison} \label{sec:comparison}
In order to give a comparison with some related works, we reimplemented the studies~\cite{Nguyen2016} and~\cite{Bauckhage2009} which respectively require skeletons and frontal-view silhouettes as the inputs. Let us describe briefly the two methods. The researchers in~\cite{Nguyen2016} performed the task of abnormal gait detection based on a sequence of skeletons. An ensemble of hand-crafted features was proposed to describe each skeleton based on geometric angles estimated from 3D joints of the lower body. Each feature vector was then converted into a codeword using a $k$-means clustering. A HMM was employed to model the change of walking postures during a walking cycle. The gait index was finally estimated for each gait cycle as the resulting likelihood. The work~\cite{Bauckhage2009} also performed the task of feature extraction for each instant posture based on the binary silhouette instead of skeleton. This stage was done with the support of a dynamic lattice. The temporal factor was considered by a simple concatenation of the feature vectors extracted from consecutive frames. A binary Support Vector Machine (SVM) was used to classify normal and abnormal gaits.

In our reimplementation, we built a HMM for~\cite{Nguyen2016} and a one-class SVM for~\cite{Bauckhage2009} since our study is considered as a unsupervised learning problem. This consideration is reasonable because in practical situations, there is a very high number of possible abnormal walking gaits. Therefore, a supervised learning may not provide an efficient model with a high level of generalization. The experimental results of these three models are presented in Table~\ref{table:comparison}.
\begin{table*}[t]
\centering
\caption{Experimental results of related works.}
\label{table:comparison}
\vspace{5pt}
\begin{tabular}{l||lccc}
\hline
\multirow{2}{*}{\textbf{Model}} & \multirow{2}{*}{\textbf{Data type}} & \multicolumn{3}{c}{\textbf{Classification error}}\\ \cline{3-5} 
 & & per-frame & per-segment & per-sequence \\ \hline \hline
HMM~\cite{Nguyen2016} & skeleton & - & 0.335 & 0.250 \\ \hline
One-class SVM~\cite{Bauckhage2009} & silhouette & 0.399 & 0.227 & 0.139 \\ \hline
Ours (non-weighted sum) & skeleton & 0.316 & 0.284 & 0.278 \\
Ours (weighted sum) & skeleton & \textbf{0.230} & \textbf{0.188} & \textbf{0.139} \\ \hline
\end{tabular}
\end{table*}

Since each gait sequence (1200 frames) contains only one gait type, we just focus on the per-sequence results. Our index estimation that used a weighted sum gave the best ability of classification. It also demonstrated that the raw features that are automatically extracted by our auto-encoder can give a good description of walking gaits.

\section{Conclusion} \label{sec:conclusion}
This paper proposes a method that estimates the gait abnormality index based on a sequence of skeletons determined by a Kinect 2. Instead of requiring hand-crafted features as some related works, we employed auto-encoders to perform this task automatically. Our system consists of 3 auto-encoders, in which the input of each one is an array of joint coordinates along the corresponding axis in 3D space. Each model can be considered as a weak gait index estimator. By proposing a weighted sum of such weak indices, the estimation ability of our system significantly improves the outputted index. Our experiments showed that our index is accurate for describing different walking gaits compared with some related works. In further works, Procrustes analysis is an interesting alternative to the coordinate normalization. Besides, the correlation between per-coordinate auto-encoders may be considered to improve the index estimator.

\subsubsection*{Acknowledgment}
The authors would like to thank the NSERC (Natural Sciences and Engineering Research Council of Canada) for having supported this work
. This research is funded by Funds for Science and Technology Development of the University of Danang under project ``Research and Proposing a Method Measuring Human Gait Abnormality Index based on Skeleton''.

\bibliography{references}
\bibliographystyle{iclr2019_conference}

\end{document}